*Original Article*

# A Dynamic Model for Bus Arrival Time Estimation based on Spatial Patterns using Machine Learning

B. P. Ashwini[1], R. Sumathi[2], H. S. Sudhira[3]

[1,2]*Department of Computer Science and Engineering, Siddaganga Institute of Technology, Karnataka, India*
[3] *Gubbi labs, LLP, Karnataka, India*

[1]*Corresponding Author : ashvinibp@sit.ac.in*

***Abstract -*** *The notion of smart cities is being adapted globally to provide a better quality of living. A smart city's smart mobility component focuses on providing smooth and safe commuting for its residents and promotes eco-friendly and sustainable alternatives such as public transit (bus). Among several smart applications, a system that provides up-to-the-minute information like bus arrival, travel duration, schedule, etc., improves the reliability of public transit services. Still, this application needs live information on traffic flow, accidents, events, and the location of the buses. Most cities lack the infrastructure to provide these data. In this context, a bus arrival prediction model is proposed for forecasting the arrival time using limited data sets. The location data of public transit buses and spatial characteristics are used for the study. One of the routes of Tumakuru city service, Tumakuru, India, is selected and divided into two spatial patterns: sections with intersections and sections without intersections. The machine learning model XGBoost is modeled for both spatial patterns individually. A model to dynamically predict bus arrival time is developed using the preceding trip information and the machine learning model to estimate the arrival time at a downstream bus stop. The performance of models is compared based on the R-squared values of the predictions made, and the proposed model established superior results. It is suggested to predict bus arrival in the study area. The proposed model can also be extended to other similar cities with limited traffic-related infrastructure.*

## 1. Introduction

The notion of smart cities [1] [2] has facilitated institutions to exploit the potential of ICT in transforming cities to provide a better quality of life. The smart cities' smart mobility component [3] [4] primarily focuses on ensuring smooth and safe commuting between any two locations for the citizens. The mobility component also promotes the public transit system [5] as an eco-friendly alternative for private modes and emphasizes using ICT [6] in this direction. The recent developments in the Intelligent Transportation System (ITS) [7] [8] [9], Advanced Transportation System(ATS), and Advanced Passenger Information System (APIS) [10] aim at providing better LoS to commuters. One of the aims of the APIS is to assist the commuters by facilitating up-to-the-minute information about the public transit system and aid the commuters in taking decisions on their mode, route, etc.,

Several existing works on the Bus Travel and Arrival Time (BTAT) prediction are based on statistical, dynamic, time series, machine learning, and deep learning models [11]. Conventionally, travel time prediction is done on two scales [12]; long-term, where the travel time is predicted 1-2 days ahead of the current time, and short-term, where predictions are made in 1-2 hours from the current time. The precision of the predictions mainly depends on the data sources available [13]; the static data such as road geometry, bus stops and intersections count, etc., along with the dynamic traffic flow information in the network plays a vital role. The availability of such datasets drives the accuracy of the prediction models. Several cities lack the infrastructure to capture live traffic flow information across the entire network. Installing the required infrastructure is mammoth and economically not viable as various past research has relied on the automatic vehicle location of buses and other vehicles, probe vehicle data, RFID at fixed locations, etc.

In this context, this work proposes a novel hybrid model to predict the Bus Arrival Time (BAT) in the short term using only the location data of the public transit buses. The proposed model combines the predictions of the extreme gradient boosting (XGBoost) machine learning technique and an analytical model developed using the preceding Bus Travel Time (BTT) and the spatial characteristics of the sections. The location data of the city service buses of Tumakuru is used for the study.

The remaining parts of the article are presented as follows. The existing literature on BTAT is reviewed and discussed in section 2. The proposed model for BTAT prediction using the limited dataset is illustrated in section 3. In section 4, the results and the discussions are presented. Finally, in section 5, the work is summarized as a conclusion.

## 2. Literature review

The recent developments in ICT and computing technology have leveraged the available resources to improve the quality of living across the globe. Travel/bus arrival time prediction is an important topic researched in the recent past, where the location data of the public transit buses are the most common dataset [13] used.

Authors [14] propose an approach to combine a radial basis neural network model and an online adjustment model. The online adjustment model adjusts the network model results based on the traffic's dynamic situation. To prove the validity, the model was deployed on a study route in Dalian city of China. The proposed model outperformed individual models such as linear regression, backpropagation, and radial basis function network models. In [15], a comparison of models such as historic average, Artificial Neural Network (ANN), and Kalman Filter(KF) for predicting the BTT is conducted. Authors considered robustness and accuracy of predictions as the metrics to compare the models and found that ANN outperforms the other models. The authors also suggested that these models can be used to implement the Advanced Public Transport System. Travel time prediction for multiple bus routes is proposed by [16] for routes in Shenzhen, China. The authors propose a hybrid model combining the baseline predictions of Support Vector Machines with the dynamic model KF.

Authors [17] have proposed a model to predict the travel time of buses in the short term in a dynamic manner using the historical data and the current information of the vehicle for which the prediction is made. A clustering algorithm: k-medoids, is used to extract the profiles in the historical data. The authors suggest that the proposed model may be used in the context of intercity trips. A performance comparison of the k-NN, KF, and Autoregressive Integrated Moving Average (ARIMA) models for dedicated and mixed traffic lanes is conducted [18]. The authors compared the model's individual and ensemble in predicting the arrival and travel time of buses in three cities in India. The dedicated lane route showed stability in prediction across models, whereas the combination of the mixed lanes k-NN-KF technique demonstrated the best results. A speed-based travel/arrival time prediction is proposed in [19]. The experiments were conducted in Samara, Russia. The proposed model was compared with Linear Regression (LR), basic ANN models, and Support Vector Regression (SVR), to highlight that it can be used for the implementation of prediction systems in the future.

Similarly, authors in [20] have predicted the arrival time of buses using limited datasets, compared the LR and SVR models, and found that SVR performs better than LR. Authors in [21] have studied a route in Chennai, India, with high variance and have applied SVR to predict the BTT. The spatial-temporal SVRs developed demonstrated improvement in the predictions compared to the existing model deployed. In a study [22] conducted in Charlotte, North Carolina, travel times are predicted for short horizons analyzed the performance of four machine learning models Regression Trees, XGBoost, Long Short-Term Memory (LSTM) neural network, and Random Forests (RF). The performance was assessed based on prediction accuracy and reliability, and RF showed better performance in both accuracy and reliability. Authors in [23] have highlighted the importance of a deep understanding of the spatial-temporal features through Recurrent Neural Networks. They have proposed a wide and deep model for travel time prediction. The experiments were conducted using the datasets of two big cities in China, and the model proposed demonstrated good results. Authors in [24] have compared the suitability of linear and non-linear machine learning models for BTT prediction. They have concluded that for tier-2 cities like Tumakuru, non-linear models are suitable, and ensemble tree models such as RF and Gradient Boosting Regression Trees(GBRT) are best for schedule preparation and travel time prediction.

Overall, it is concluded from the existing literature that there are no generic models for accurate BTAT prediction, and most of the solutions are location specific. With the advancements in ICT and ITS, there are several cities in which these technologies are yet to be implemented. In this direction, this work aims to predict BTAT in the study area with limited traffic-related infrastructure. Hybrid and machine learning models have been widely used in BTAT recently; hence a suitable machine learning model is identified from the previous study, and a hybrid model is proposed.

## 3. Data

The location data of the Tumakuru city service buses of March 2021 is used to develop the machine learning and bus arrival estimation model. 600 trips of the route, Tumkur Bus stand to Kyathasandra is selected; also, the other routes sharing the common sections with the study route are identified for modeling. The sample location data is given in table 1, and the study route map with the bus stops is given in Figure 1. The information on the study route augmented with the spatial characteristics is given in table 2.

Table 1. Sample location data of public transit bus

| Vehicle No | Date and Time | Latitude | Longitude | Odometer | Speed |
|---|---|---|---|---|---|
| KA-06-F-0836 | 01-03-2021 07:33:37 | 13.34286 | 77.09886 | 6121.51 | 0 |
| KA-06-F-0836 | 01-03-2021 07:33:44 | 13.34292 | 77.09894 | 6121.52 | 6 |
| KA-06-F-0836 | 01-03-2021 07:33:54 | 13.34298 | 77.09917 | 6121.54 | 6 |
| KA-06-F-0836 | 01-03-2021 07:34:04 | 13.34314 | 77.09919 | 6121.56 | 7 |
| KA-06-F-0836 | 01-03-2021 07:34:14 | 13.34328 | 77.09897 | 6121.6 | 12 |
| KA-06-F-0836 | 01-03-2021 07:34:24 | 13.34313 | 77.09872 | 6121.63 | 12 |
| KA-06-F-0836 | 01-03-2021 07:34:34 | 13.34271 | 77.09856 | 6121.68 | 21 |
| KA-06-F-0836 | 01-03-2021 07:34:44 | 13.3422 | 77.09839 | 6121.74 | 17 |
| KA-06-F-0836 | 01-03-2021 07:34:54 | 13.3419 | 77.09829 | 6121.78 | 0 |
| KA-06-F-0836 | 01-03-2021 07:34:58 | 13.34181 | 77.09826 | 6121.79 | 8 |
| KA-06-F-0836 | 01-03-2021 07:35:08 | 13.34144 | 77.09814 | 6121.83 | 19 |
| KA-06-F-0836 | 01-03-2021 07:35:18 | 13.34085 | 77.09798 | 6121.9 | 26 |
| KA-06-F-0836 | 01-03-2021 07:35:28 | 13.3403 | 77.09781 | 6121.96 | 17 |
| KA-06-F-0836 | 01-03-2021 07:35:38 | 13.33983 | 77.09767 | 6122.01 | 22 |
| KA-06-F-0836 | 01-03-2021 07:35:48 | 13.33932 | 77.09753 | 6122.07 | 15 |
| KA-06-F-0836 | 01-03-2021 07:35:58 | 13.33908 | 77.09748 | 6122.1 | 9 |

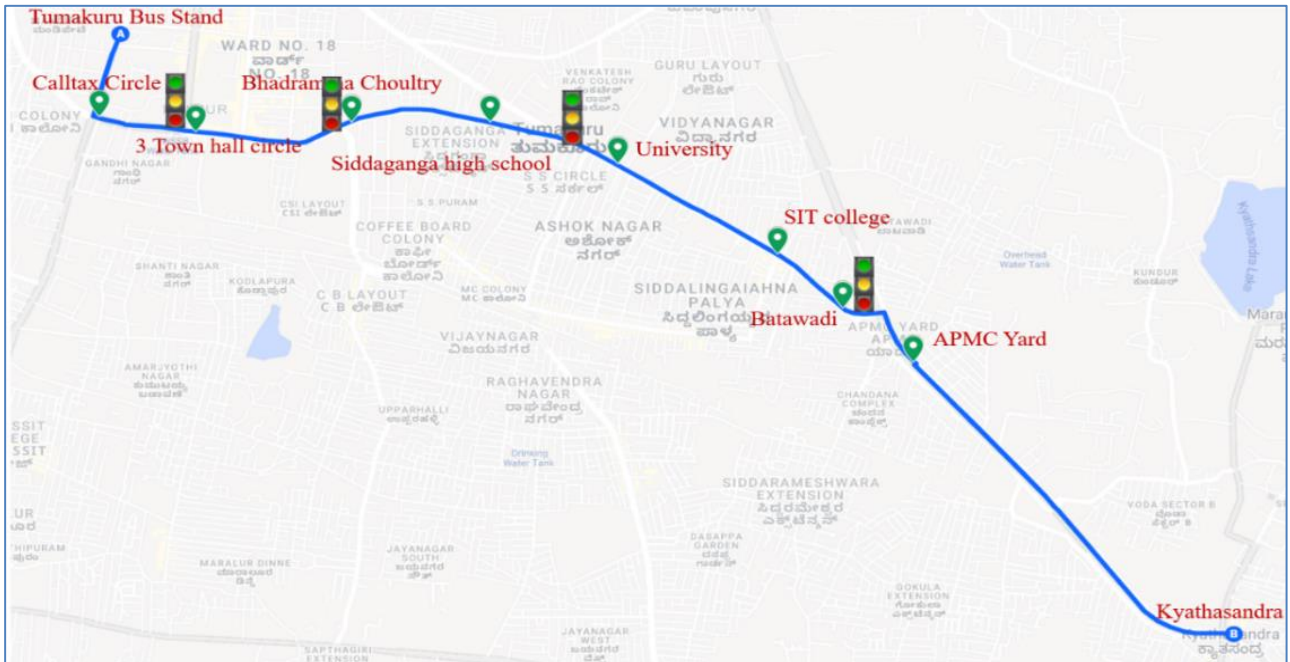

**Fig. 1 The route map of the Tumkur Bus stand to Kyathasandra with bus stops**

Table 2. Route information

| Section | Start and end bus stops | Length (meters) | LUP | Intersection |
|---|---|---|---|---|
| 1 | Tumkur Bus Stand(TBS) - CallTax Circle (CTC) | 600 | CBD | No |
| 2 | CallTax Circle – Town Hall Circle (THC) | 450 | CBD | Yes |
| 3 | Town Hall Circle - Bhadramma Choultry (BC) | 750 | CBD | Yes |
| 4 | Bhadramma Choultry - Siddagnaga High School (SHS) | 650 | IC | No |
| 5 | Siddagnaga High School – Government University (GU) | 550 | IC | Yes |
| 6 | University- SIT College (SIT) | 1000 | ISU | No |
| 7 | SIT College – Batawadi (BT) | 350 | ISU | Yes |
| 8 | Batawadi - APMC Yard (APMC) | 400 | OSU | No |
| 9 | APMC Yard – Kyathasandra (KY) | 2200 | OSU | No |

CBD: Central Business District, IC: Inner City, ISU: Inner Suburban, OSU: Outer Suburban

## 4. Methods
### 4.1 Preprocessing
Variability analysis of the travel times in the selected route (up and downstream) is presented in [25]. The same procedure is followed to pre-process the data in the current article. The raw location data are pre-processed to remove missing and erroneous entries and are transformed into aggregates at the bus stop level for further study. The travel time between bus stops depends on factors such as Land Use patterns (LUP), signalized intersections, un-signalized intersections, section length, etc. As an additional feature, LUP and the length of the section are augmented to the trip aggregates. Unsignalized intersections are not included in the study. The presence of signalized intersections has a major influence on travel time; hence the trip aggregates are split as sections (bus stop to bus stop) with and without intersections. Sections 2,3,5, and 7 are Sections with InterSections (SIS) and sections 1,4,6,8 and 9 are the sections without intersections and are called Normal Sections (NS) in the future in this article.

### 4.2. Extreme Gradient Boosting (XGBoost)
XGBoost is a tree model used in applications that need classification and regression. It is an efficient implementation of the gradient boosting regression tree technique with boosting approach that ensembles multiple simple regression trees. The trees are built based on the residuals at each level to optimize the loss function. It is a widely used model to predict BTT [26][27][28]. The scikit-learn [29], a free machine learning library, is used to implement XGBoost. According to the previous study [24], the non-linear models are best suited for BTT prediction in the current study area. Therefore, the XGBoost technique is used to forecast the BTT. The XGBoost is trained for trip aggregates data with intersections and without intersections separately, and individual models are developed. A few parameters and its corresponding values in the study are given in table 3.

### 4.3. Bus Arrival Time Estimation Model
The real-time information on the traffic flow in the link, congestion alerts, and occurrence of incidents are unavailable in the study area. The data available is limited to only the location data of the buses. Hence, the preceding buses are considered probe vehicles, and the travel time of the preceding buses is assumed to reflect the delays in the link. The latest travel time information about the sections, such as travel time and running speed, are extracted from preceding buses of the same and other routes (multi-routes). The trips that have preceding trips with less than 30 minutes difference from the current trip start time are considered for the study. Based on the spatial patterns, the sections of the route are divided as follows:

- Sections without intersections (Normal Section (NS))

- Sections with intersections (SIS)

A dynamic BAT Estimation Model (BATEM) is proposed in this work that predicts the BAT at a downstream bus stop based on the preceding trip data of buses and the machine learning model predictions for both spatial patterns. The implementation flow of the proposed model is given in figure 2. The list of symbols used in the article is summarized in table 4.

The proposed BATEM estimates the BAT for individual bus stops by considering the current time as the start time of the section. The BAT for each section j is estimated as given in equation 1, in which $C_{time}$ is the current time and the $ATT_t^j$ is the Adjusted Travel Time estimated for section j at start time t.

$$BAT_j = C_{time} + ATT_t^j \quad (1)$$

Adjusted Travel Time is the time estimated based on the type of the section using equation (2). It is the weighted average of the Forecasted Travel Time $FTT_t^j$ by the XGBoost, and the estimated travel time based on preceding trip travel time $PTT_i^j$. The weights are computed as given in equation (5).

Table 3. Few important parameters of the XGBoost Regression technique used

| Parameters | Values Used | Description |
|---|---|---|
| objective | squared error | A loss function that estimates the distance of a predicted value from the actual regressed value |
| alpha | 1.00 | L1 regularization term on weights |
| learning_rate | 0.05 | Step size used in each iteration during optimizing a loss function |
| n_estimators | 200 | Number of trees in the additive model |
| colsample_bytree | 0.60 | A subsample ratio of columns during the construction of each tree. Subsampling is done for every constructed tree |
| max_depth | 3.00 | The upper bound for the number of nodes in the weak learner regression tree |

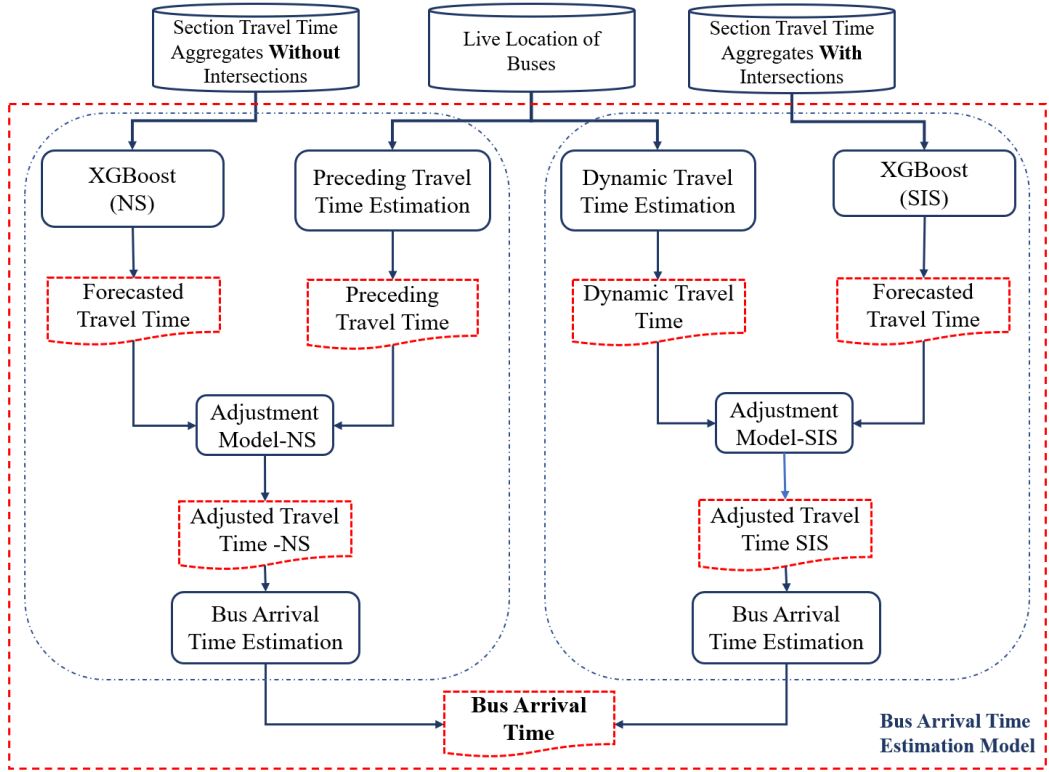

**Fig. 2 Flow of the proposed model**

$$ATT_t^j = \begin{pmatrix} w_1 * FTT_t^j + w_2 * PTT_i^j \mid for\ NS \\ w_1 * FTT_t^j + w_2 * DTT_i^j \mid for\ SIS \end{pmatrix} \quad (2)$$

For normal sections, the $ATT_t^j$ is estimated using the $FTT_t^j$ and preceding trip travel time $PTT_i^j$, whereas for sections with intersections, the estimations are made using Dynamic Travel Time $DTT_i^j$ and $FTT_t^j$. The $PTT_i^j$ is the travel time of a preceding bus for the section with start time i, where (t - i) is less than 30 minutes. If multiple bus information is available, then the minimum of (t - i) is considered. and the $DTT_i^j$ is estimated using the running time of the preceding trip $PRT_i^j$ Along with the standard dwell time $DWD_j$ And average delay at intersections $ID_j$ as given in equation (3).

$$DTT_i^j = PRT_i^j + DWD_j + ID_j \quad (3)$$

The equation to estimate the preceding trip running time is presented in equation (4), where $d_j$ is the distance of section j, and $RS_i^j$ is the running speed of each section of the preceding trip.

$$PRT_i^j = \frac{d_j}{RS_i^j} \quad (4)$$

The weights in equation (2) are estimated by equation (5). These weights are derived by estimating the $x_1$; the coefficient of determination (R-squared) of the machine learning model and $x_2$ The historical data shows the correlation of the current trip to that of the preceding trip.

$$w_1 = \frac{x_1}{x_1+x_2} \quad w_2 = \frac{x_2}{x_1+x_2} \quad (5)$$

The R-squared (R2) is a statistical measure determining the strength of the linear relationship between the parameters under consideration. The formula for computing the coefficient of determination is given in equation 6 for $n$ trips. Where $p_{y,j}$ is the dependent variable (actual travel time of trip y for section j), the function $f(T_{z,j})$ returns the predicted travel time using the independent variables, the $T_{z,j}$ is a table containing the independent variables day-of-the-week, the trip's start time, section number, and the land use pattern. $\mu p_{y,j}$ is the mean of the actual travel time in test data.

$$x_1 = 1 - \sum_{y=1}^{n}(p_{y,j} - f(T_{z,j}))^2 \Big/ \sum_{i=1}^{n}(p_{y,j} - \mu p_{y,j})^2 \quad (6)$$

Analyzing the correlation [30] is a method used to measure the level of deviation in one feature due to deviation in another. The correlation analysis of the preceding trip travel and the current trip travel time in the archived data is conducted to derive the correlation coefficients. Preceding trips traversing the same sections within 30 minutes to the current time are considered. The standard equation for computing the correlation is given in equation 7, where $x_2$ is the correlation coefficient, $tt_c$ is the current trip travel time and $tt_p$ is the travel time of the preceding trip. The $\mu tt_c$ and $\mu tt_p$ are the average of the current and preceding trip travel times.

$$x_2 = \frac{\Sigma\left((tt_c - \mu tt_c) * (tt_p - \mu tt_p)\right)}{\sqrt{\Sigma(tt_c - \mu tt_c) * \Sigma(tt_p - \mu tt_p)}} \quad (7)$$

All the equations are applied to sections with intersections and without intersections to dynamically predict the BAT at a downstream bus stop using only the live and historic location data of the buses.

## 5. Results and Discussion

The traffic's stochastic behaviour sometimes decreases the machine learning model's performance. The live data regarding traffic flow, incidents, and delays influence the travel duration and, thus, the BAT. With limited information, such as only live and historical location data of the city service buses, predicting the arrival time of the buses is a colossal task. Hence, this article presents a hybrid mode of estimating the arrival time of the buses, where the machine learning model and a model based on the preceding trips are combined to predict the BTAT. The estimated values for the parameters in the proposed model are tabulated in table 5. The average intersection delay for each intersection is estimated in [34].

The R-squared (x1) value indicates that the sections without intersections are more predictable by the machine learning model than those with intersections. The intersections in Tumakuru city operate in clearance mode with no bus priority [32] [33]. The length of sections with intersections is less than 800 meters. Thus, the variations in the waiting delay at signalized intersections affect the travel time. Hence the sections having intersections demonstrate more variations and less predictability.

A plot of the actual travel time against the forecasted travel by the XGBoost model and the travel time estimated based on the proposed model of a few sample trips is presented in Figure 3 and Figure 4. Figure 3 depicts the predictions of the sections with intersections, and figure 4 of the sections without intersections. The start time for each section and the travel time in seconds is plotted on the X-axis and the Y-axis, respectively. It is evident from the plot that the proposed model can recognize the dynamic changes in the traffic flow and estimate the travel time better as compared to the XGBoost model.

Table 5. The computed values of the constants used in the study

| Spatial level | $w_1$ | $w_2$ | $x_1$ | $x_2$ | Intersection delay in seconds (section) |
|---|---|---|---|---|---|
| SIS | 0.45 | 0.55 | 0.40 | 0.50 | 47 (2), 45 (3), 57 (5), 20 (7) |
| NS | 0.56 | 0.44 | 0.71 | 0.55 | --------- |

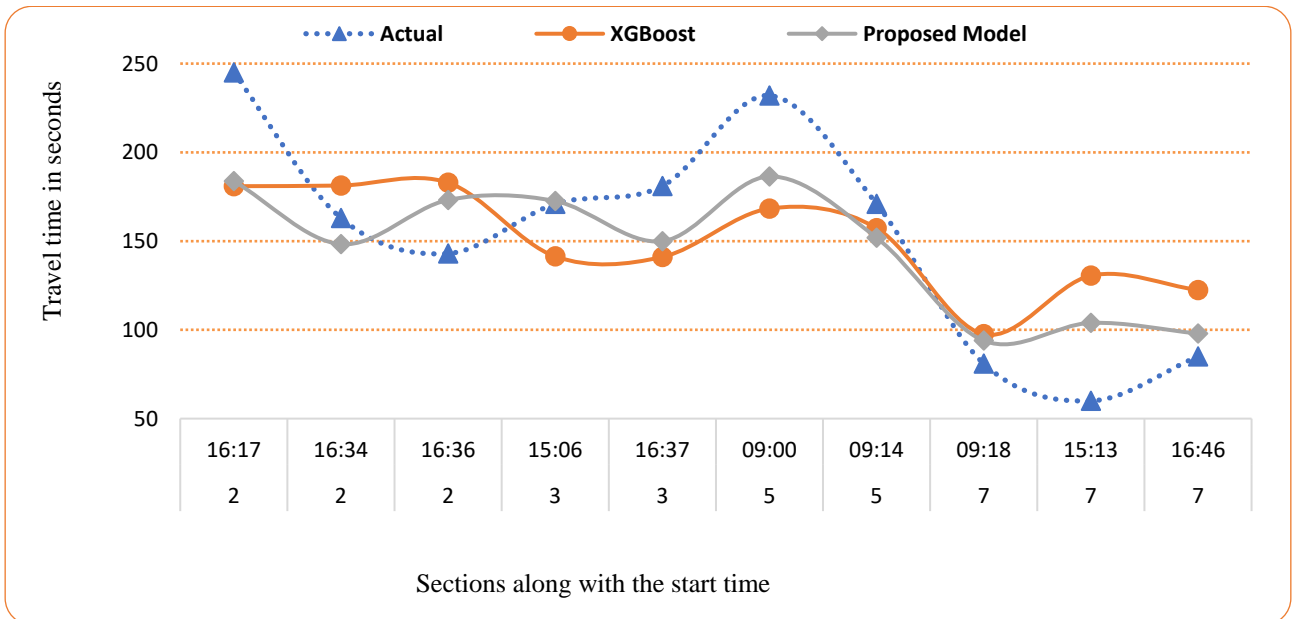

**Fig. 3 Actual vs predicted travel times of XGBoost and proposed models for SIS**

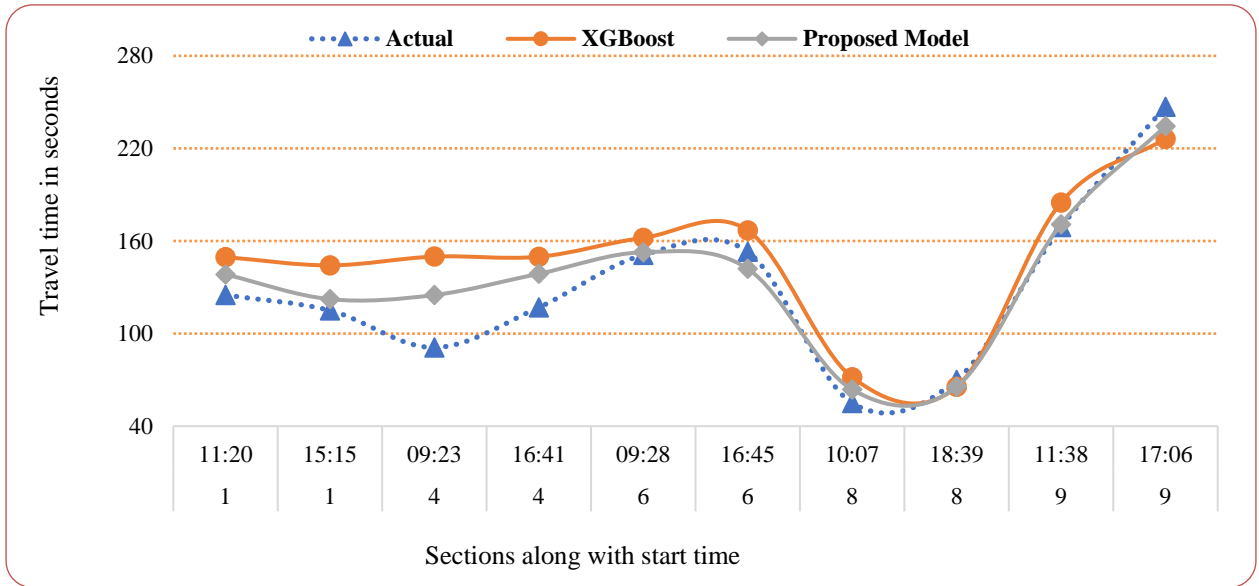

**Fig. 4 Actual vs predicted travel times of XGBoost and proposed models for NS**

The predictions made for one trip on the 11th of March 2021, starting at 15:14:00, are plotted to depict the predictions of the XGBoost and the proposed model in figure 5. The actual travel time of the trip is 1209 seconds, the total duration predicted by XGBoost is 1339 seconds, and the proposed model is 1261. The XGBoost predicted individual sections' travel time with a mean error of 30 seconds, whereas the proposed models' average error was 18 seconds. It implies that the travel time predicted by the proposed model is better than the machine learning model.

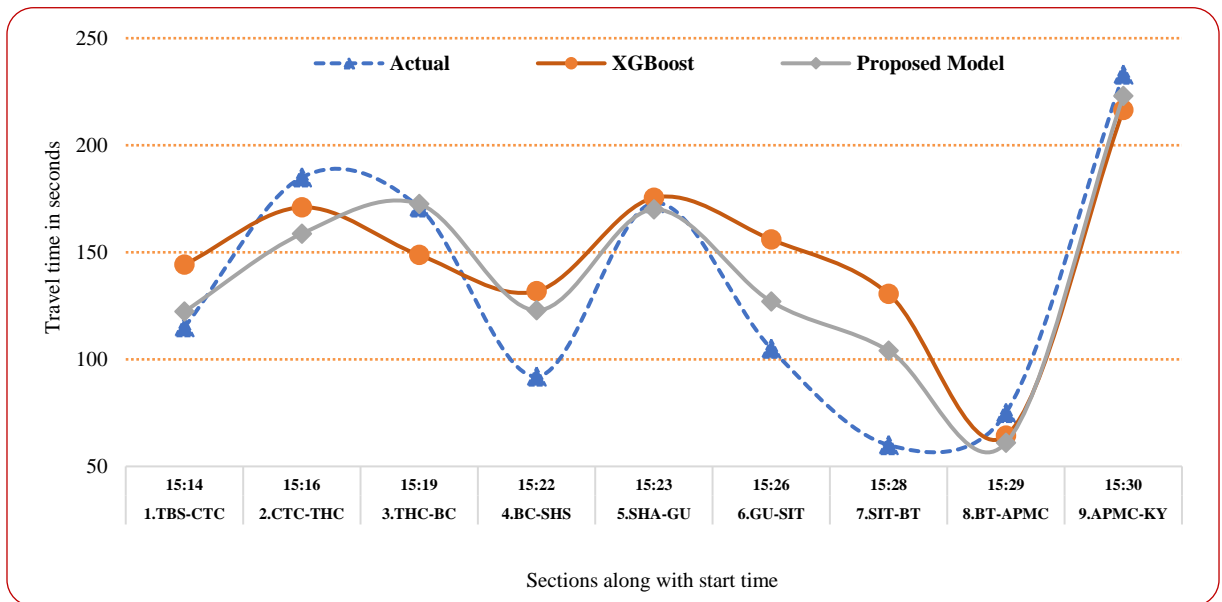

**Fig. 5 Actual vs predicted travel times of XGBoost and proposed models for a sample trip**

The R-squared of the XGBoost model is 0.4 and 0.71. In contrast, the R-squared of the proposed model is 0.59 and 0.80 for sections with and without intersections, respectively, demonstrating significant improvement in the predictions. The performance values are plotted in Fig. 6.

The data available in the study area is limited to the locations of the public transit buses. With such a constraint on the data, where the live information of the traffic, congestion, or incidents is unavailable, the proposed model has attempted to predict the BAT. The model's performance suggests that the proposed model is suitable for predicting

BAT in the study area. And there are several similar cities with limited traffic infrastructure for which the proposed model can be extended. To further improve the performance of the proposed model, the current trip information, dynamic traffic flow information of the links, and details of incidents, if any, are also vital as the up-to-the-minute information reduces the uncertainties of the prediction.

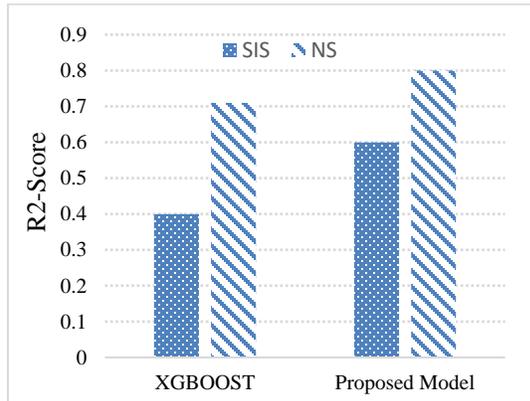

**Fig. 6 Performance comparison of the proposed model with XGBoost**

## 6. Conclusion

A dynamic model to predict bus arrival time using only limited data sets based on machine learning is proposed in this article. Based on the previous study results, the base travel time is predicted using the XGBoost model. The proposed Bus Arrival Time Estimation Model estimates the bus arrival time for individual bus stops for a selected route in Tumakuru, India. The sections between the bus stops were divided into two discrete spatial patterns: a section with intersections and sections without intersections. Each spatial pattern was modeled separately using the XGBoost and the proposed model. The proposed model estimated the bus arrival time based on both machine learning model results and the preceding trip travel time and demonstrated better accuracy than the XGBoost model for both spatial patterns. The preceding trip within 30 minutes to that of the current demonstrated a positive correlation, illustrating its usefulness in forecasting travel times. The proposed model exhibited better accuracy based on the R-squared values. The proposed model can be suggested for bus arrival time forecasting in the study route and extended to other routes to predict the city-wide bus arrivals. Similar cities can employ the proposed model as well. If available, the up-to-the-minute details of the traffic flow, incidents, and other delays can further improve the forecasting. The model can be updated in the future once the up-to-the-minute data stream is available. Overall, with the available limited datasets, the proposed model is a promising option for forecasting the bus arrival time.

## Acknowledgements

The authors wish to thank Smart City Limited of Tumakuru (TSCL) for providing the required logs of the city service buses.